\documentclass[letterpaper, 10 pt, conference]{ieeeconf}
\usepackage{amsmath,amsfonts}
\usepackage{array}
\usepackage{enumerate}
\usepackage[utf8]{inputenc}
\usepackage[T1]{fontenc}
\usepackage{amsfonts}
\usepackage{amssymb}
\IEEEoverridecommandlockouts 
\usepackage{algorithm}  
\usepackage{algpseudocode}  
\usepackage{amsmath}  
\usepackage{breqn}
\usepackage{textcomp}
\usepackage{stfloats}
\usepackage{url}
\usepackage{verbatim}
\usepackage{graphicx}
\usepackage{cite}
\usepackage{subcaption}
\usepackage[font=footnotesize]{caption}

\usepackage{hyperref}
\hypersetup{
    colorlinks=true,
    linkcolor=blue,
    filecolor=magenta,      
    urlcolor=cyan,
    }
\title{\LARGE \bf
NaviSTAR: Socially Aware Robot Navigation with Hybrid Spatio-Temporal Graph Transformer and Preference Learning}

\author{Weizheng Wang$^{1}$, Ruiqi Wang$^{1}$, Le Mao$^{2}$, and Byung-Cheol Min$^{1}$ 
\thanks{$^{1}$SMART Laboratory, Department of Computer and Information Technology, Purdue University, West Lafayette, IN, USA. {\tt\small{[wang5716 ,wang5357,minb]@purdue.edu}.}}
\thanks{$^{2}$College of Mechanical and Electrical Engineering, Beijing University of Chemical Technology, Beijing, China. \tt\small{2020030286@buct.edu.cn}.}}

\begin{document}

\maketitle
\begin{abstract}
Developing robotic technologies for use in human society requires ensuring the safety of robots' navigation behaviors while adhering to pedestrians' expectations and social norms. However, understanding complex human-robot interactions (HRI) to infer potential cooperation and response among robots and pedestrians for cooperative collision avoidance is challenging. To address these challenges, we propose a novel socially-aware navigation benchmark called NaviSTAR, which utilizes a hybrid Spatio-Temporal grAph tRansformer to understand interactions in human-rich environments fusing crowd multi-modal dynamic features. We leverage an off-policy reinforcement learning algorithm with preference learning to train a policy and a reward function network with supervisor guidance. Additionally, we design a social score function to evaluate the overall performance of social navigation. To compare, we train and test our algorithm with other state-of-the-art methods in both simulator and real-world scenarios independently. Our results show that NaviSTAR outperforms previous methods with outstanding performance\footnote{The source code and experiment videos of this work are available at: \url{https://sites.google.com/view/san-navistar} \label{2}}.
\end{abstract}
\section{Introduction}

Recent advances in machine intelligence have led to the integration of robots into human daily life. Service robots are now used in homes, and delivery robots are deployed in towns, among other applications. This integration requires robots to perform socially aware navigation in spaces shared with humans. Specifically, when navigating in a human-filled environment, robots must not only avoid collisions but also exhibit socially compliant manner, regulating their movements to create and maintain a pleasant spatial interaction experience for other pedestrians \cite{garrell2012cooperative}.

Existing works on socially aware robot navigation can generally be classified into two mainstreams: \textit{decoupled} and \textit{coupled} approaches. Decoupled approaches involve building a model that encodes crowd dynamics to forecast the uncertainty of pedestrians' movements, including intentional paths of humans. These forecasts then act as additional parameters for traditional collision-free path planners to generate optimal trajectories \cite{du2011robot,bennewitz2005learning}. However, such methods tend to overlook latent human-robot interactions (HRI), leading to uncertainty explosions, particularly in highly complex environments. Additionally, issues like freezing robot problems and reciprocal dance problems can arise \cite{trautman2015robot}.

\begin{figure}[!t]
\centering
\includegraphics[width=0.80\columnwidth]{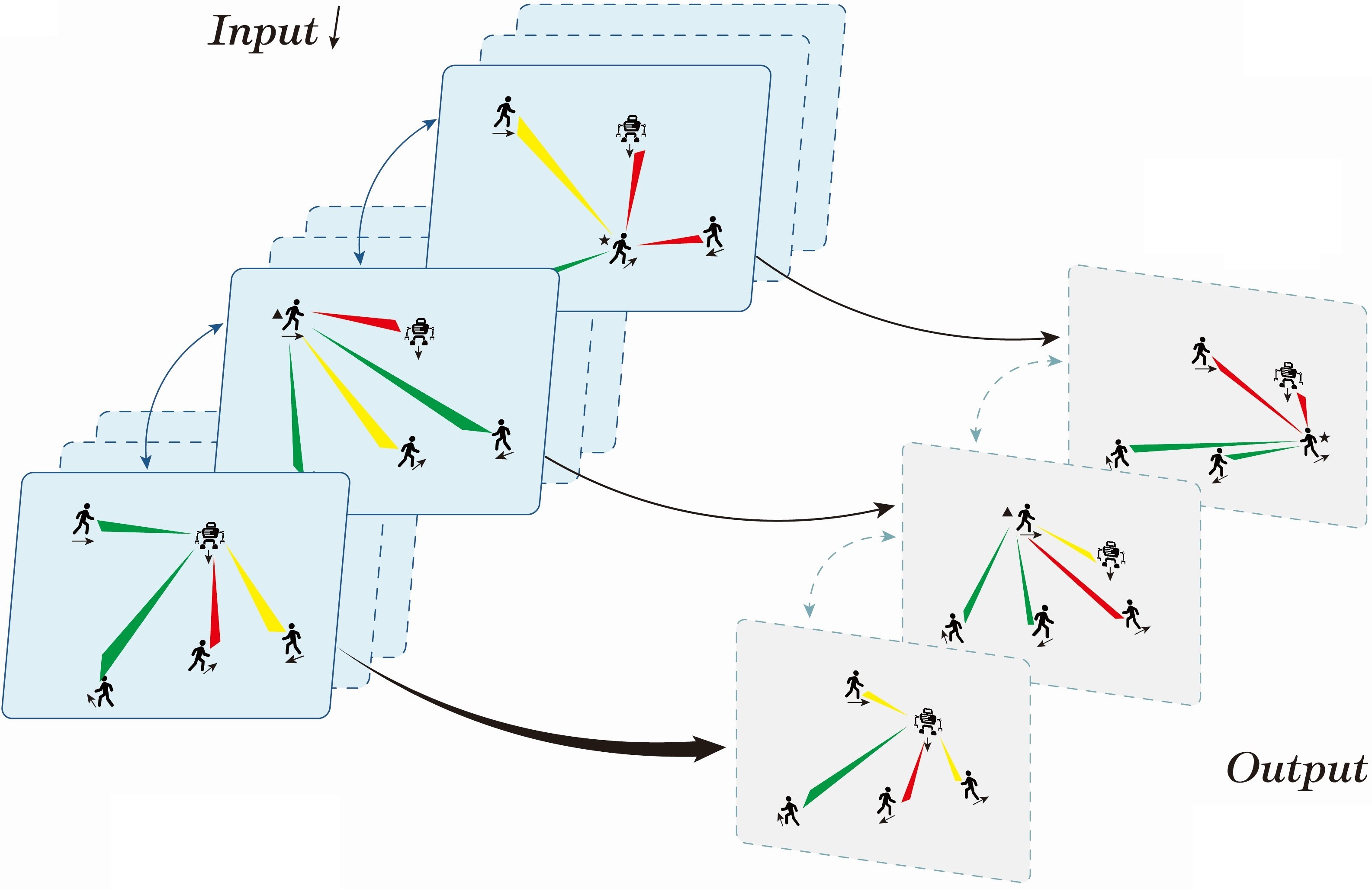}
\vspace{-4pt}
\caption{HRI \& HHI illustration: NaviSTAR uses a hybrid spatio-temporal graph transformer to capture complex environmental dynamics, which aggregates all the spatial and temporal attention maps from each agent.}
\vspace{-15pt}
\label{fig:STG}
\end{figure}
As alternatives, coupled approaches have archived better performance by treating the socially aware navigation problem as a cooperative collision avoidance task \cite{kretzschmar2016socially,chen2017decentralized,chen2019crowd,chen2020relational,liu2021decentralized,SunM-RSS-21,wang2022feedback}. These approaches train neural networks of robots to implicitly infer HRI state representation, demonstrating human-level intelligence. The HRI state representation is then fed through a policy network to learn socially acceptable navigation actions based on Reinforcement Learning (RL) approaches.

Despite the promising results of these methods, several limitations still need to be addressed. For instance, interaction state representation learning involves highly complex interactions between crowds and robots, including human-human interactions (HHI) and HRI, each with spatial and temporal features. Moreover, decision-making is not solely driven by these interactions, but also by the latent dependencies across them. However, previous studies have either failed to fully consider the aforementioned interactions in both spatial and temporal dimensions \cite{chen2017decentralized,chen2019crowd,chen2020relational,liu2021decentralized}, or they have struggled to efficiently capture the hidden dependencies across multi-modal interactions \cite{liu2022socially}. On the other hand, for policy learning, most studies tend to rely on a handcrafted reward function, which is difficult to quantify broad social compliance and can lead to the reward exploitation problem during robot policy optimization. While the most recent study \cite{wang2022feedback} 
has shown that active preference learning can lead to more preferred and natural robot actions compared to handcrafted reward functions.

To address the aforementioned gaps, we propose NaviSTAR, a \textbf{S}patio-\textbf{T}emporal gr\textbf{A}ph t\textbf{R}ansformer-based approach with preference learning. Specifically, we introduce a hybrid spatio-temporal graph transformer framework to encode the latent dependencies across HHI and HRI in both spatial and temporal dimensions into the interaction state representation (Fig. \ref{fig:STG}). Then, we combine off-policy RL with preference learning to encode human intelligence and expectations into the robotic reward function neural network.

The main contributions of this paper can be summarized as: 1) We introduce a fully connected spatio-temporal graph to model HHI and HRI via both spatial and temporal dimensions. Within this graph, a hybrid spatio-temporal graph transformer is developed to capture and fuse heterogeneous latent dependencies across different aspects, into the state representation of environmental dynamics; 2) We leverage off-policy learning and preference learning to train a robot policy that considers social norms and human expectations; and 3) We conducted extensive simulation experiments and a real-world user study to demonstrate the benefits of our model.
\vspace{-4pt}
\section{Background}
\subsection{Related Works}
Deep reinforcement learning based approaches that treat pedestrians as agents with individual policies to unfold a multi-agent system have been shown to achieve better performance in socially aware robot navigation. In optimizing robot paths for social compliance, the RL framework consists of two parts: 1) a state representation learning network, and 2) a policy learning network. The first part implicitly learns the interactions and cooperation among agents and the following part optimizes a robot navigation policy based on the learned state representation. To encode agent interactions, many research efforts have been conducted. For example, \cite{chen2017decentralized} introduced a pair-wise navigation algorithm to present the interaction between a pair-wise human and robot, then \cite{chen2019crowd} developed an attention mechanism to cover crowd interactions inside. Furthermore, \cite{chen2020relational} used a graph convolution network to improve the presentation of HRI as a spatial graph and predict navigation strategies by model-based RL. However, such three studies overlook the temporal features of HHI and HRI. More recently, \cite{liu2021decentralized,liu2022socially} created a graph of social environments by borrowing temporal edges to comprehensively describe the potential correlation of a multi-agent system. While this method integrates both spatial and temporal features of HHI and HRI, it fails to sufficiently model and coalesce latent spatio-temporal dependencies.

On the other hand, fewer studies have been conducted to improve the model from the policy learning perspective. Most of them either rely on handcrafted reward function or inverse RL for policy learning. However, the handcrafted reward functions are hard to quantify complex and broad social norms and can result in the reward exploitation problem that leads to unnatural robot behaviors \cite{wang2022feedback}. Moreover, while inverse RL can introduce human expectations to robot policy through demonstrations, it suffers from expensive and inaccurate demonstrations as well as extensive feature engineering. To address these limitations, \cite{wang2022feedback} proposed an active preference learning to tailor a reward model by human feedback, leading to more preferred and natural robot actions.

\subsection{Spatio-Temporal Graph and Multi Modal Transformer}
The spatio-temporal graph (ST-graph) is a conditional random field that captures high-level semantic dependencies among objects, in which nodes are objects and edges link two nodes to present spatial or temporal interaction between node pairs. For example, \cite{ref23} predicted pedestrian trajectories with an ST-graph, \cite{li2022bevformer} used an ST-graph for Bird's-Eye-View maps generation, and \cite{chen2022bidirectional} utilized an ST-graph for traffic flow forecasting. These works demonstrate the powerful feature representation of ST-graphs and the success of transformer-based sequence learning in time series forecasting. And the model of pedestrians' movements is highly spatial-temporal feature-dependent, which is not only rely on surrounding obstacles distribution or spatial interaction, but also can be predicted from trajectory temporal dependencies.

To fuse cross-modal features, a multi-modal transformer is introduced to bridge the gaps between heterogeneous modalities. For example, \cite{tsai2019multimodal} designed a multi-modal transformer to align human speech from vision, language, and audio modalities, while \cite{chen2021multimodal} formulated a multi-modal transformer in computational pathology to learn the mapping between medical images and genomic features. Recently, \cite{wang2022husformer} proposed a human state prediction framework using a multi-modal transformer from multiple sensors data.

Inspired by these works, we adapt a spatio-temporal graph transformer to capture long-term dependencies in socially aware navigation tasks, which describe the spatial-temporal human-robot interaction. We then aggregate the spatial-temporal dependencies into a multi-modal transformer network to estimate potential cooperation and collision avoidance, considering the multimodality and uncertainty of pedestrians' movements.

\section{Methodology}
\subsection{Preliminaries}

Following previous works \cite{chen2017decentralized,chen2019crowd,liu2021decentralized}, we define the socially aware navigation task as a partially observable Markov decision process (POMDP) represented by a tuple $\langle \mathcal S,\mathcal A,\mathcal{O},\mathcal{\hat{P}} ,\mathcal R,\gamma,n, \mathcal S_0 \rangle$, where $n$ is the number of agents. $\rm \mathbf s^{jn}_{\rm t} \in \mathcal{S}$ represents the joint state of MDP at timestep t, with $\rm \mathbf{s}^{jn}_{\rm{t}} = [\rm \mathbf{s}^{robot}_{\rm{t}},\rm \mathbf{s}^{ho}_{\rm{t}}]$, and $\rm \mathbf{s}^{ho}_{\rm{t}}$ are the observable states of humans. For each agent, the state is $\rm \mathbf{s}_{t} = [\rm \mathbf{s}^{o}_{\rm{t}},\rm \mathbf{s}^{uo}_{\rm{t}}]$, where $ {\rm \mathbf{s}^{o}_{\rm{t}}} = [ p_{\rm x}, p_{\rm y}, v_{\rm{x}}, v_{\rm{y}},radius] \in \mathcal{O}$ represents the observable state, which includes position, velocity, and radius. The hidden state $\mathbf{s}^{\rm{uo}}_{\rm{t}}=[ {p}_{\rm{x}}, {p}_{\rm{y}},v_{\rm{pref}},{\theta}]$ contains the agent's self-goal, preferred speed, heading angle. The $v_{\rm{pref}}$ is used as a normalization term in the discount factor for numerical reasons from \cite{chen2017decentralized}. The robot action $\rm \mathbf{a}_{\rm{t}} \in \mathcal{A}$, and $\mathcal{\hat{P}}$ represents the state transition of the environment. The discount factor $\gamma$ is in the range [0,1]. $\mathcal R$ is the reward function, which is represented by a network with parameter $\alpha$ as $\tilde{r}_{\alpha}$ in our algorithm. The initial state distribution is $\mathcal S_O$. Finally, the objective $\mathcal{J}$ is defined as follows:
\begin{equation}
\mathcal{J}_{(\pi)} = \mathbb{E}_{(a,s)}[\sum_{t} \gamma^{{t} {\cdot} {v_{pref}}}\cdot  \tilde{r}_{\alpha}\left(\mathbf{s}_{\rm{t}}, \mathbf{a}_{\rm{t}}\right)]
\end{equation}

\begin{figure*}[!t]
\centering
\includegraphics[width=0.99\linewidth]{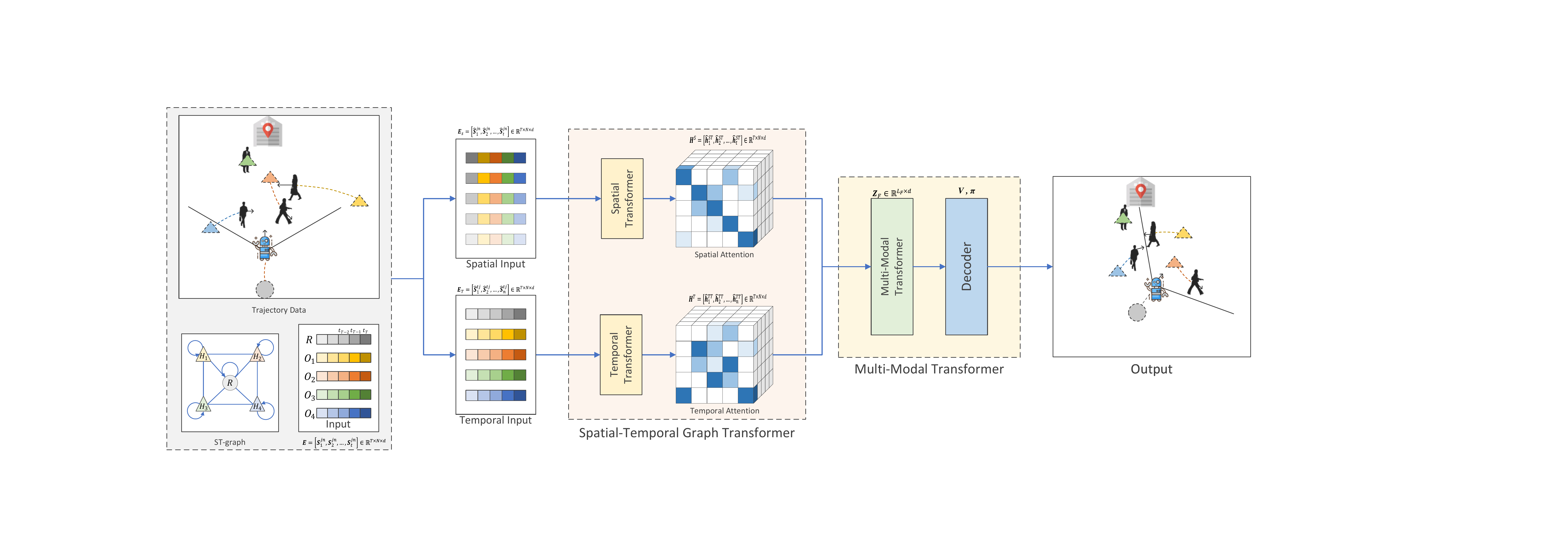}
\vspace{2pt}
\caption{NaviSTAR architecture: The social navigation planner utilizes a spatial-temporal graph transformer block and a multi-modal transformer block to abstract environmental dynamics and human-robot interactions into an ST-graph for path planning in crowd-filled environments. The spatial transformer is designed to capture both static and dynamic spatial interactions and to generate spatial attention maps, while the temporal transformer presents long-term temporal dependencies and creates temporal attention maps. The multi-modal transformer is deployed to adapt the uncertainty and multimodality of crowd movements, aggregating all heterogeneous spatial and temporal features. Finally, the planner performs the next timestep action by a decoder.}
\vspace{-10pt}
\label{fig:architecture}
\end{figure*}

\vspace{-10pt}
\subsection{Spatio-Temporal Graph Representation}
\label{HEL}
We represent the socially aware navigation scenario as a ST-graph $\mathcal{G}=(\mathcal{N}, \mathcal{E}, \mathbf{M})$, where $\mathcal{N}$ is the set of nodes representing agents. The robot policy and observed human states are factorized into robot nodes and human nodes. $\mathcal{E}_{\rm{S}}$ represents the spatial edge, parameterized by the spatial transformer parameter matrix $\mathbf{M}_{\rm{S}}$, while $\mathcal{E}_{\rm{T}}$ represents the temporal edge, parameterized by the temporal transformer parameter martix $\mathbf{M}_{\rm{T}}$. The adjacency matrix $\mathbf{M} \in \mathbb R^{\rm N \times N}$ is constructed using a Gaussian kernel. The environmental representation function $\mathcal{F}$ is learned from a cross-hybrid spatio-temporal graph transformer. Once the environmental dynamics is constructed using the ST-graph, the robotic HRI state representation $\mathbf Z_{\rm{F}}$ can be calculated. Subsequently, $\mathbf Z_{\rm{F}}$ is deconstructed as an action vector $\mathbf{a}_{\rm{t+1}}$ to drive the robot, accomplished by a decoder.
\begin{equation}
    \mathbf Z_{\rm{F}} = \mathcal F (\mathbf s^{\rm jn}_1,\cdots,\mathbf s^{\rm jn}_{\rm{t}};\mathcal G)
\end{equation}



\subsection{NaviSTAR Architecture}

\noindent\textbf{Overview:} As shown in Fig.~\ref{fig:architecture}, we develop a transformer-based framework to construct environmental dynamics representation using a ST-graph. First, the environment state ${\mathbf{E}}=[\mathbf {s}^{\rm jn}_1,\cdot\cdot\cdot,\mathbf{s}^{\rm jn}_{\rm t}]$ is fed as input into the network, which include all agents' observed states from the robot's field of view (FOV) during timestep 1 to timestep t. Then, a spatial-temporal graph transformer network is introduced to capture long-term spatial and temporal dependencies, generating spatial attention maps and temporal attention maps (Fig. \ref{fig:attention map}). Moreover, in order to fuse above cross-modality interactions features, a multi-modal transformer is adapted to capture the multimodality of crowd navigation environments by aggregating all spatial attention maps and temporal attention maps from each agent (as shown in Fig. \ref{fig:attention map}). Lastly, the robot obtains the value ${\rm{V}}_{\mathbf{s}_{\rm{t}}}$ and the policy $\pi_{(\mathbf{a}_{\rm{t}}|\mathbf{s}_{\rm{t}})}$ from a decoder to navigate.

\begin{figure*}[!t]
\centering
\includegraphics[width=0.95\linewidth]{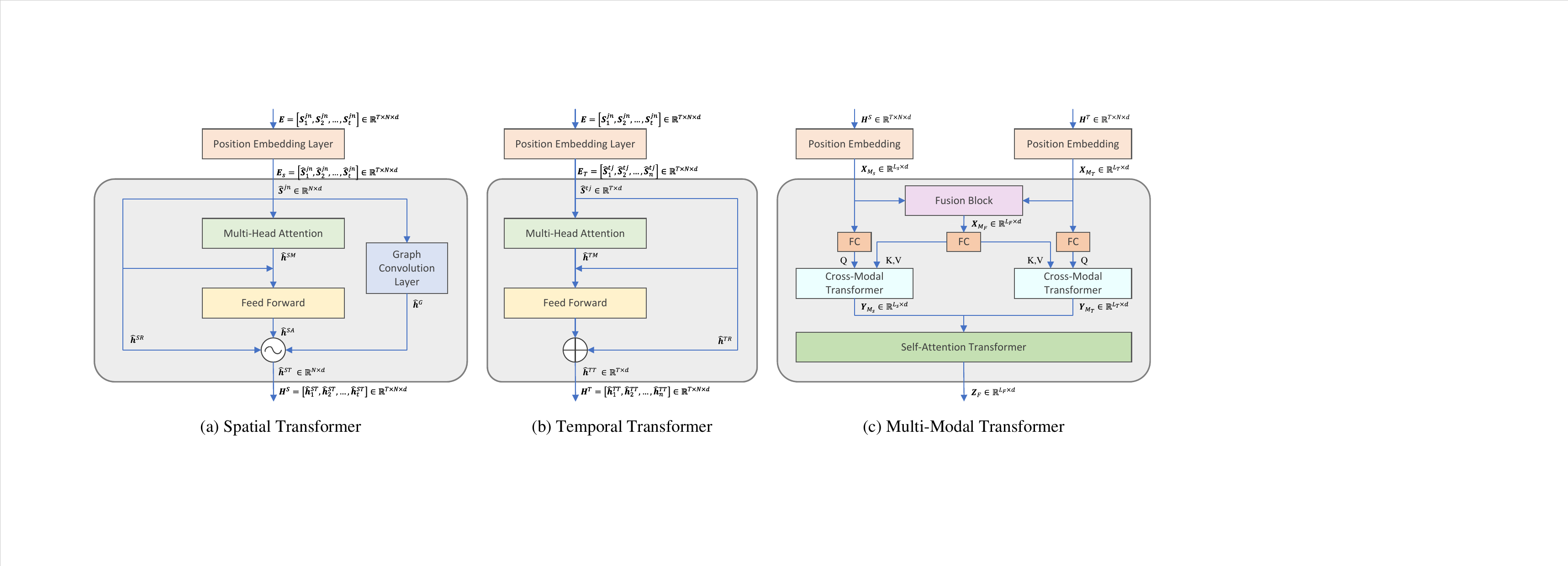}
\vspace{-3pt}
\caption{NaviSTAR neural network framework: (a) Spatial Transformer leverages a multi-head attention layer and a graph convolution network along the time-dimension to represent spatial attention features and spatial relational features; (b) Temporal Transformer utilizes multi-head attention layers to capture each individual agent's long-term temporal attention dependencies; and (c) Multi-Modal Transformer fuses heterogeneous spatial and temporal features via a multi-head cross-modal transformer block and a self-transformer block to abstract the uncertainty of multimodality crowd movements.}
\vspace{-10pt}
\label{fig:network detail}
\end{figure*}

\begin{figure}[!t]
\centering
\vspace{4pt}
\includegraphics[width=0.98\columnwidth]{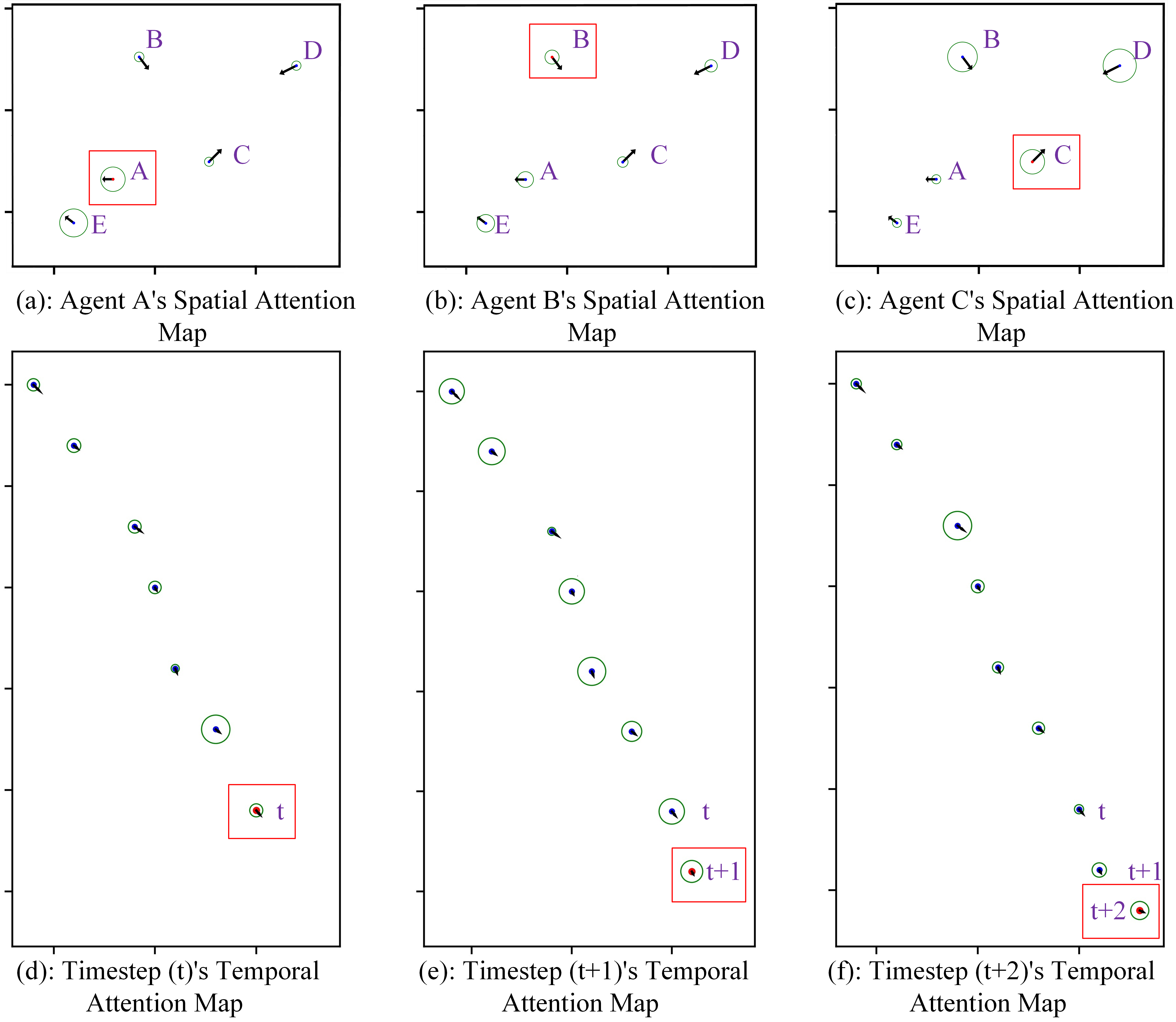}
\vspace{-3pt}
\caption{An illustration of spatial and temporal attention maps: sub-figures (a), (b), and (c) exhibit the spatial attention maps from different agents at the same timestep; (d), (e), and (f) present the temporal attention maps from different timesteps in the same agent's view. The radius of the circle represents the importance level with respect to the agent in the red square.}
\vspace{-20pt}
\label{fig:attention map}
\end{figure}

\noindent\textbf{Spatial Transformer Block:} The spatial-temporal transformer block calculates the spatial attention and temporal attention separately from the environment state $\mathbf{E}$, which is used to represent the edges of spatio-temporal graph. In the spatial transformer, we rotate the input matrix to align each agent's state with the same timestep to calculate the spatial dependencies. This block encodes the spatial attentions map (representing the importance of neighboring agents) and spatial relational features (capturing the distance and relative orientation among pairwise agents) by the multi-head attention block and graph convolution layer. As shown in Fig.~\ref{fig:network detail}(a) the environment state $\mathrm{\mathbf{E}} \in \mathbb R^{\rm T\times \rm N\times \rm d}$ is fed into a fully connected layer to embed the positional information based on \cite{vaswani2017attention} to compute the spatial embeddings $\mathrm{\mathbf{E}}_{\rm s}=[\mathbf {\hat{s}}^{\rm jn}_1,\cdot\cdot\cdot,\mathbf {\hat{s}}^{\rm jn}_{\rm t}]$. Similarly, the temporal embeddings $\mathrm{\mathbf{E}}_{\rm t}=[\mathbf {\hat{s}}^{\rm tj}_1,\cdot\cdot\cdot,\mathbf {\hat{s}}^{\rm tj}_{\rm n}]$ are aligned by a fully connected layer in the temporal transformer block, as shown in Fig.~\ref{fig:network detail}(b). The spatial embeddings of each timestep $\mathbf{\hat{s}}^{\rm jn} \in \mathbb R^{\rm N\times \rm d}$ are then used as input for the spatial transformer.

In order to capture the spatial correlational dependencies in each timestep, we introduce a graph convolution neural network (GCN) based on Chebyshev polynomial approximation\cite{kipf2017semisupervised}. This GCN is used to learn the static spatial correlation features $\mathbf{\hat{h}}^{\rm{G}} \in \mathbb R^{\rm N\times \rm d}$ among multiple agents.
\vspace{-5pt}
\begin{equation}
\mathbf{\hat{h}}^{\rm{G}} = {\sum_{\rm k=0}^{\rm{K}} {\theta_{\rm{k}}} \rm{\mathbf{T}_k(\mathbf{\hat L})\hat {\mathbf{s}}^{jn}}}
\end{equation}
\noindent where $\mathbf{I}_n, \mathbf{D}$ is an identity matrix and the diagonal degree matrix of the adjacency matrix $\mathbf{M}$. The symmetric normalized Laplacian matrix is then calculated as $\mathbf{L} = \mathbf{I}_n - \mathbf{D}^{(-\frac{1}{2})} \mathbf{M} \mathbf{D}^{(\frac{1}{2})}$. After that, the scaled Laplacian matrix is defined as $ {\hat {\mathbf{L}}} =2 {\mathbf L} / {\lambda_{\max }}-\mathbf{I}_{n}$, where ${\lambda_{\max }}$ is the largest eigenvalue of $\mathbf{L}$. The parameter $\theta$ is a weighted parameter, and $\rm{K}$ is the kernel size of the graph convolution. Finally, the result $\rm \hat{\mathbf{h}}^{\rm{G}}$ is calculated by Chebyshev polynomials $\rm \mathbf{T}_k$.


Even though GCN can capture static spatial relational features, the robot needs to understand potential dynamic spatial features with respect to adjacent agents' intents and attentions. Thus, we utilize multi-head attention mechanism to generate each agents' spatial attention maps, as shown in Fig.~\ref{fig:attention map}, to capture the dynamic spatial dependencies 
$\mathbf{\hat{h}}^{\rm{SM}}$.
\begin{equation}    
\operatorname{Atten}\left(\rm \mathbf Q_{\rm{ST}}^{(i)},\rm \mathbf K_{\rm{ST}}^{(i)},\rm \mathbf V_{\rm{ST}}^{(i)}\right)={\mathrm{softmax} (\frac{ \rm \mathbf Q_{\rm{ST}}^{(i)} \rm \mathbf ({K_{\rm{ST}}^{(i)}})^{\top}}{\sqrt{ \rm{d}_{{h}}}}) \cdot \rm \mathbf V_{\rm{ST}}^{(i)}}   
\end{equation}
\begin{equation}
    \begin{aligned}
        \operatorname{Multi}\left(\mathbf Q_{\rm{ST}}^{(\rm i)},\mathbf K_{\rm{ST}}^{(\rm i)},\mathbf V_{\rm{ST}}^{(\rm i)}\right)&=f_{\rm{fc}}( \rm{head}_1,\cdots, \rm{head}_{\rm{h}} ) = \mathbf{\hat{h}}^{\rm{SM}},\\
         \rm{head}_{(\cdot)}&=\operatorname{Atten}_{(\cdot)}\left(\rm \mathbf{Q}_{\rm{ST}}^{(i)}, \rm \mathbf{K}_{\rm{ST}}^{(i)}, \rm \mathbf{V}_{\rm{ST}}^{(i)}\right)
    \end{aligned}
\end{equation}

\noindent where $f_{\rm{fc}}()$ is a linear function, and $\rm h$ is the number of heads in the multi-head attention mechanism, $\rm i$ is the label of timesteps in the spatial transformer, while it represents the label of pedestrians in the temporal transformer. $\mathbf Q_{\rm{ST}}^{\rm(i)}, \mathbf K_{\rm{ST}}^{\rm(i)}, \mathbf V_{\rm{ST}}^{\rm(i)}$ are query, key and value matrices with dimension $\rm{d_h}$, which are computed from $\mathbf{\hat{s}}^{\rm{jn}}$ or $\mathbf{\hat{s}}^{\rm{tj}}$ in the spatial transformer or temporal transformer block. They are generated using a fully connected layer with respect to parameter matrices $\{\mathbf{M}_{\rm{S}},\mathbf{M}_{\rm{T}}\}$.


We use the rectified linear unit (ReLU) activation function to capture non-linear features in the feedforward network. Additionally, residual connections are added to the spatial and temporal transformer networks to  accelerate convergence and stabilize the framework\cite{he2016deep}. 
\begin{equation}
\mathbf{\hat{h}}^{\rm{SR}} = \mathbf{\hat{h}}^{\rm{SM}} + \mathbf{\hat{s}}^{\rm{jn}}
\vspace{-3pt}
\end{equation}
\begin{equation}
\mathbf{\hat{h}}^{\rm{SA}} = \mathbf{\hat{h}}^{\rm{SR}} + \rm{ReLU}(\rm{ReLU}(\mathbf{\hat{h}}^{\rm{SR}} {\mathbf W_{S}^{(1)}})\mathbf W_{S}^{(2)})\mathbf W_{S}^{(3)}
\vspace{-3pt}
\end{equation}
\noindent where $\rm \mathbf W_{S}^{(1)},\rm \mathbf W_{S}^{(2)},\rm \mathbf W_{S}^{(3)}$ are  the weight matrices  of the hidden layers in the feedforward network.

Lastly, a gate mechanism $\rm{d}$ is used to combine the representation of spatial features $\mathbf{\hat{h}}^{\rm{ST}}$,
\begin{equation}
     {\rm{d}} = \mathrm{sigmoid}\left[  f_{\rm{fc}} (\mathbf{\hat{h}}^{\rm{SA}})+f_{\rm{fc}}(\mathbf{\hat{h}}^{\rm{G}})  \right]
\end{equation}
\begin{equation}
    {\mathbf{h}}^{\rm{ST}}= \rm{d}\cdot (\mathbf{\hat{h}}^{\rm{SA}})+(1-\rm{d})\cdot (\mathbf{\hat{h}}^{\rm{G}})
\end{equation}
Then, we pack up the final spatial features $\mathbf{\hat{h}}^{\rm{ST}}$ in a spatial attention matrix $\mathbf{H}^{\rm S} = [\mathbf{\hat{h}}^{\rm{ST}}_{1},\cdot\cdot\cdot,\mathbf{\hat{h}}^{\rm{ST}}_{\rm{t}}] \in \mathbb R^{\rm T\times \rm N\times \rm d}$.

\noindent\textbf{Temporal Transformer Block:} Due to the highly motion-dependency of the temporal dimension, we designed temporal edges connecting each agent's context pair-wise across timesteps. For example, since motion is continuous, we can predict future velocity and position based on current trajectory history. The temporal transformer block is parallel to spatial transformer block in our framework. And each agent's individual trajectory is aligned as a row vector of input. As shown in Fig.~\ref{fig:network detail}(b), the temporal transformer block is similar to the spatial transformer, but without a GCN. Firstly, the temporal embeddings ${\mathbf{{E}_t}}$ are fed into a multi head attention layer to capture temporal features $\mathbf{\hat{h}}^{\rm{TM}}$ and create temporal attention maps for each agents as shown in Fig.~\ref{fig:attention map} from each agent's trajectory data. Next, a feedforward network and residual connection structure are deployed in the framework to obtain the final temporal dependencies $\mathbf{\hat{h}}^{\rm{TT}}$, which is same as the spatial transformer.
\begin{equation}
\mathbf{\hat{h}}^{\rm{TR}} = \mathbf{\hat{h}}^{\rm{TM}} + \mathbf{\hat{s}}^{\rm{tj}}
\vspace{-3pt}
\end{equation}
\begin{equation}
\vspace{-3pt}
\mathbf{\hat{h}}^{\rm{TT}} = \mathbf{\hat{h}}^{\rm{TR}} + \rm{ReLU}(\rm{ReLU}(\mathbf{\hat{h}}^{\rm{TR}} {\mathbf W_{T}^{(1)}})\mathbf W_{T}^{(2)})\mathbf W_{T}^{(3)}
\vspace{1pt}
\end{equation}
\noindent where $\rm \mathbf W_{T}^{(1)},\rm \mathbf W_{T}^{(2)},\rm \mathbf W_{T}^{(3)}$ are  the weight matrices  of the hidden layers in the feedforward network. Finally, we package the temporal feature $\mathbf{\hat{h}}^{\rm{TT}}$ into a temporal attention matrix $\mathbf{\rm{{\mathbf H}^T}} = [\mathbf{\hat{h}}^{\rm{TT}}_{1},\cdot\cdot\cdot,\mathbf{\hat{h}}^{\rm{TT}}_{\rm{n}}] \in \mathbb R^{\rm T\times \rm N\times \rm d}$.

\noindent\textbf{Multi-Modal Transformer Block:} In order to capture the multimodalities and uncertainty of crowd movements, we have developed a multi-head multi-modal transformer block based on \cite{wang2022husformer,tsai2019multimodal} that fuses the heterogeneous spatial and temporal attention maps of each agents and captures long-range dependencies among modalities. Once obtaining, spatial and temporal attention matrices are obtained, we utilize a multi-head cross-modal transformer to calculate the high-level representation of environmental dynamics. As shown in Fig.~\ref{fig:network detail}(c), first, the spatial and temporal feature matrices ($\rm{\mathbf H}^{S}$, $\rm{\mathbf H}^{T}$) are separately processed by a positional embedding layer from \cite{vaswani2017attention} to ensure order-invariance, resulting in $\rm{\mathbf X}_{M_{S}} \in \mathbb R^{\rm {L}_{S}\times \rm{d}}$, $\rm{\mathbf X}_{M_{T}} \in \mathbb R^{\rm {L}_{T}\times \rm{d}}$. Next, the spatial and temporal feature matrices are concatenated by a fusion block as $\rm{\mathbf X}_{M_{F}} \in \mathbb R^{\rm {L}_{F}\times \rm{d}}$ to formulate the unimodal Query $\rm{\mathbf Q_{\rm{U}}}$, fusion Key $\rm{\mathbf K_{\rm{F}}}$ and fusion Value $\rm{\mathbf V_{\rm{F}}}$ by a fully connected layer as follows:
\begin{equation}
\begin{aligned}
& \rm \mathbf Q_{\rm{U}}= \rm \mathbf X_{\rm{M}_{g}} \cdot \rm \mathbf W_{\rm{Q}_{U}} \\
& \rm \mathbf K_{\rm{F}}= \rm \mathbf X_{\rm{M}_{F}} \cdot \rm \mathbf W_{\rm{K}_{F}}\\
& \rm \mathbf V_{\rm{F}}= \rm \mathbf X_{\rm{M}_{F}} \cdot \rm \mathbf W_{\rm{V}_{F}}
\end{aligned}
\end{equation}

\noindent where $ \rm{\mathbf W}_{\rm{Q}_{U}} \in \mathbb R^{\rm d\times \rm{d}_{k}}$, $\rm{\mathbf W}_{\rm{K}_{F}} \in \mathbb R^{\rm d\times \rm{d}_{k}}$ and $\rm{\mathbf W}_{\rm{V}_{F}} \in \mathbb R^{\rm d\times \rm{d}_{v}}$ are learnable weights of the multi modal transformer.

Secondly, the unimodal feature $\rm{\mathbf X}_{M_{g}}$ is fed into multi-head cross-modal attention block with the fusion feature $\rm{\mathbf X}_{M_{F}}$ to capture high-level HRI embedding, aggregating each agent's spatial and temporal attention maps, with $\rm{g} \in \{\rm{S},\rm{T}\}$. And the $\mathbf Y^{\rm{Mul}}_{\rm{M}_{g}}$ is the output of multi-head cross-modal attention layer in the cross-modal transformer as follows:
\begin{equation}
\begin{aligned}
 \mathbf Y^{\rm{head_j}}_{\rm{M}_{g}} &= \rm{CMAtten}(\mathbf X_{\rm{M}_{g}}) \\
& = \rm{Atten}(\mathbf Q^{\rm{head_j}}_{\rm{U}}, \mathbf K^{\rm{head_j}}_{\rm{F}}, \mathbf V^{\rm{head_j}}_{\rm{F}})\\
& = \mathrm{softmax}(\frac{\mathbf Q^{\rm{head_j}}_{\rm{U}} (\mathbf K^{\rm{head_j}}_{\rm{F}})^\top}{\sqrt{\rm d_{\rm{k}}}}) \cdot \mathbf V^{\rm{head_j}}_{\rm{F}}\\
\end{aligned}
\end{equation}
\begin{equation}
\mathbf Y^{\rm{Mul}}_{\rm{M}_{g}} = {f}_{\rm{fc}}(\mathbf Y^{\rm{head_1}}_{\rm{M}_{g}}, \cdots, \mathbf Y^{\rm{head_h}}_{\rm{M}_{g}})
\end{equation}

Additionally, we incorporate a residual connection mechanism and feedforward layers into the cross-modal transformer block, based on \cite{wang2022husformer}. The output of the multi modal transformer $\mathbf Y_{\rm{M}_{g}}$ is defined as:
\begin{equation}
\mathbf Y_{\rm{M}_{g}} =  \rm{Trans}_{\rm{cross}}(\mathbf X_{\rm{M}_{g}},\mathbf Y^{Mul}_{\rm{M}_{g}})
\end{equation}

Finally, the crossed spatial feature and crossed temporal feature are adapted by an original transformer network \cite{vaswani2017attention} to further improve the efficiency of cross-modal fusion and adjust for noise. A fully-connected layer-based decoder is then designed to extract the value ${\rm{V}}_{\mathbf{s}_{\rm{t}}}$ and the policy $\pi_{(\mathbf{a}_{\rm{t}}|\mathbf{s}_{\rm{t}})}$ from the self-attention layer output $\mathbf Z_{\rm{F}} \in \mathbb R^{\rm {L}_{F}\times \rm{d}}$.
\begin{equation}
\mathbf Z_{\rm{F}} =  \rm{Trans}_{\rm{self}}(\mathbf Y_{\rm{M}_{S}},\mathbf Y_{\rm{M}_{T}})
\end{equation}


\noindent\textbf{Preference Learning and Reinforcement Learning:}

In the training procedure, we leverage preference learning \cite{lee2021bpref} and an off-policy RL algorithm, soft actor critic (SAC) \cite{haarnoja2018soft}, to implicitly encode social norms and human expectations into the reward function, policy, and value function, based on \cite{wang2022feedback}.

Preference learning aligns a reward neural network with human preferences that a human supervisor selects from different trajectory segments. Let $\varepsilon$ be a segment of a trajectory $[\mathbf{s}_{\rm{n}}^{\rm{jn}},\mathbf{a}_{\rm{n}}^{\rm{jn}},...,\mathbf{s}_{\rm{n+t}}^{\rm{jn}},\mathbf{a}_{\rm{n+t}}^{\rm{jn}}]$, and $\omega$ be a preference label from the distribution [(0,1), (1,0), (0.5,0.5)]. Firstly, two segments are displayed to the human supervisor which are sampled from the replay buffer $\mathcal B$. Then, the supervisor can choose one to update the replay buffer with human preferences, by a tuple $\langle \epsilon_{0},\epsilon_{1},\omega \rangle$, where (1,0) and (0,1) indicate left or right is better, and (0.5,0.5) represents other situations (such as nonjudgmental). The preference predictor $\mathcal{P}$ is defined as follows:
\begin{equation}
\begin{aligned}
&  {\mathcal{P}}(\alpha)\left[\varepsilon_{1}>\varepsilon_{0}\right]=\frac{\exp \sum_{\rm{t}} \tilde{r}_{\alpha}\left(\mathbf{s}_{\rm{t}}^{1}, \mathbf{a}_{\rm{t}}^{1}\right)}{\sum_{\rm{u} \in\{0,1\}} \exp \sum_{\rm{t}} \tilde{r}_{\alpha}\left(\mathbf{s}_{\rm{t}}^{\rm u}, \mathbf{a}_{\rm{t}}^{\rm u}\right)}
\end{aligned}
\end{equation}
\begin{equation}
	\omega=
	\begin{cases}
	    (1,0),& {\mathcal{P}}(\alpha)\left[\varepsilon_{0}>\varepsilon_{1}\right]>\frac{1}{2}\\
	    (0,1),& {\mathcal{P}}(\alpha)\left[\varepsilon_{1}>\varepsilon_{0}\right]>\frac{1}{2}\\
	    (0.5,0,5), & \left|{\mathcal{P}}(\alpha)\left[\varepsilon_{0}>\varepsilon_{1}\right]-{\mathcal{P}}(\alpha)\left[\varepsilon_{1}>\varepsilon_{0}\right]\right| \le 0.1
    \end{cases}
\end{equation}
\noindent where $\tilde{r}_{\alpha}$ is the learnable reward function from preferences. 

Then, a loss function $\mathcal{L}_{(\alpha)}^{R}$ is deployed to update the reward function network from the predictor with respect to real human feedback as follows:
\begin{equation}
	\begin{aligned}
		\mathcal L_{(\alpha)}^R=-\underset{\langle\varepsilon_0,\varepsilon_{1},\omega\rangle\sim \mathcal{B}}{\mathbb{E}} ( &\omega_{(1)}\log {\mathcal{P}}(\alpha)\left[\varepsilon_{0}>\varepsilon_{1}\right]\\
        +&\omega_{(2)} \log {\mathcal{P}}(\alpha)\left[\varepsilon_{1}>\varepsilon_{0}\right] )
	\end{aligned}
\end{equation}

Once the reward function $\tilde{r}_{\alpha}$ is trained, the off-policy RL algorithm SAC is used to update value function ${\rm{V}}$ and policy $\pi$, we use equations-(17),(18),(19) to develop FAPL \cite{wang2022feedback}, as our training procedure.

\section{Experiments And Results}
\subsection{Simulation Experiment}
\subsubsection{Simulation Environment}
Fig. \ref{fig33} illustrates our 2D simulator based on \cite{liu2021decentralized,chen2019crowd}. And the kinematics of environment dynamics is simplified by $\Delta \rm{t}$ from \cite{chen2017decentralized}, and human policies are developed with ORCA \cite{ORCA}. We assume that the robot is invisible to each human in the environment to optimize navigation behavior. The robot's action is defined as $\mathbf{a}_{\rm{t}}=[{v}_{\rm{x}},{v}_{\rm{y}}]$, and its FOV is set from 0° to 360°. We train and test the simulation in an open space with dimensions of ${20~m\times22~m}$, where humans are randomly generated on a circle with a radius of 8 $m$.

\begin{figure*}[htb]
\centering
\includegraphics[width=16cm]{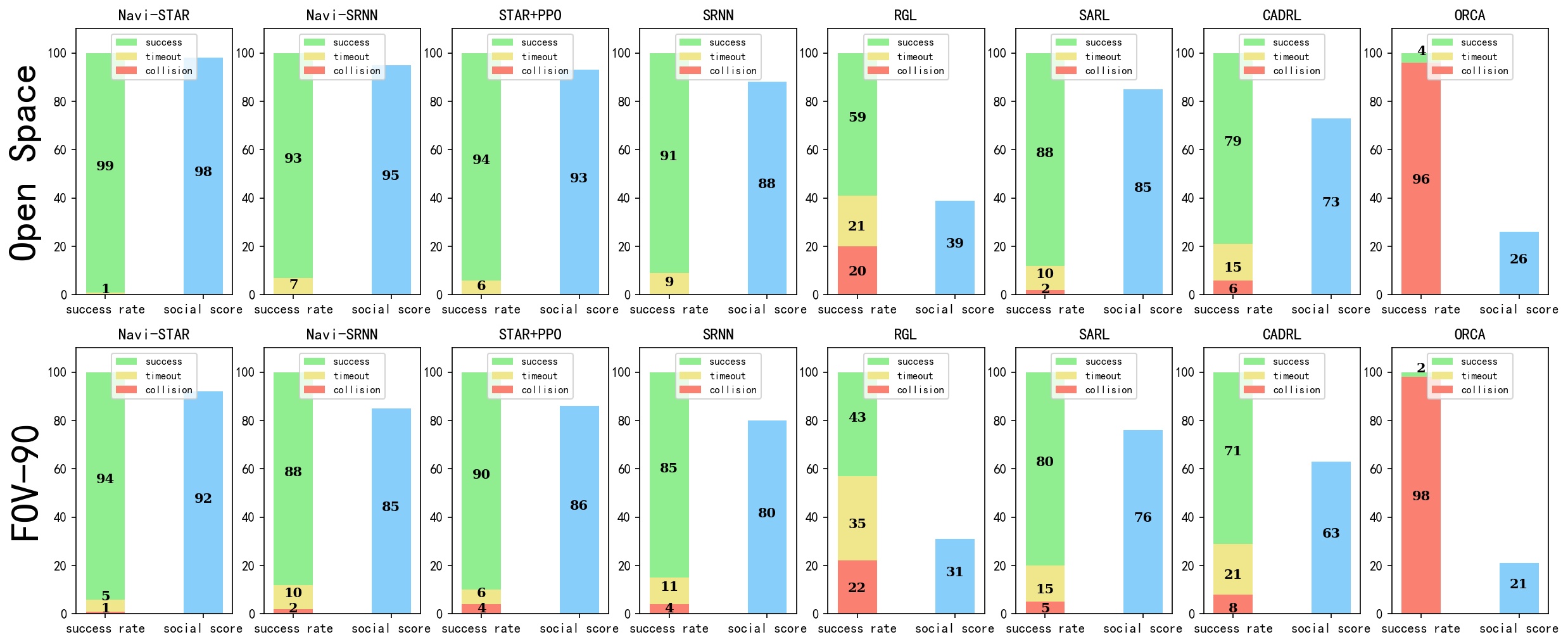}
\vspace{-1pt}
\caption{Simulation tests: the success rate (green) and social score (blue) of each policy from 500 tests with the same environment configuration (Open Space or FOV-90°). }
\vspace{-15pt}
\label{fig5}
\end{figure*}

\begin{figure}[!t]
\centering
\vspace{-5pt}
\subfloat[Open space simulator]{\includegraphics[width=3.5CM]{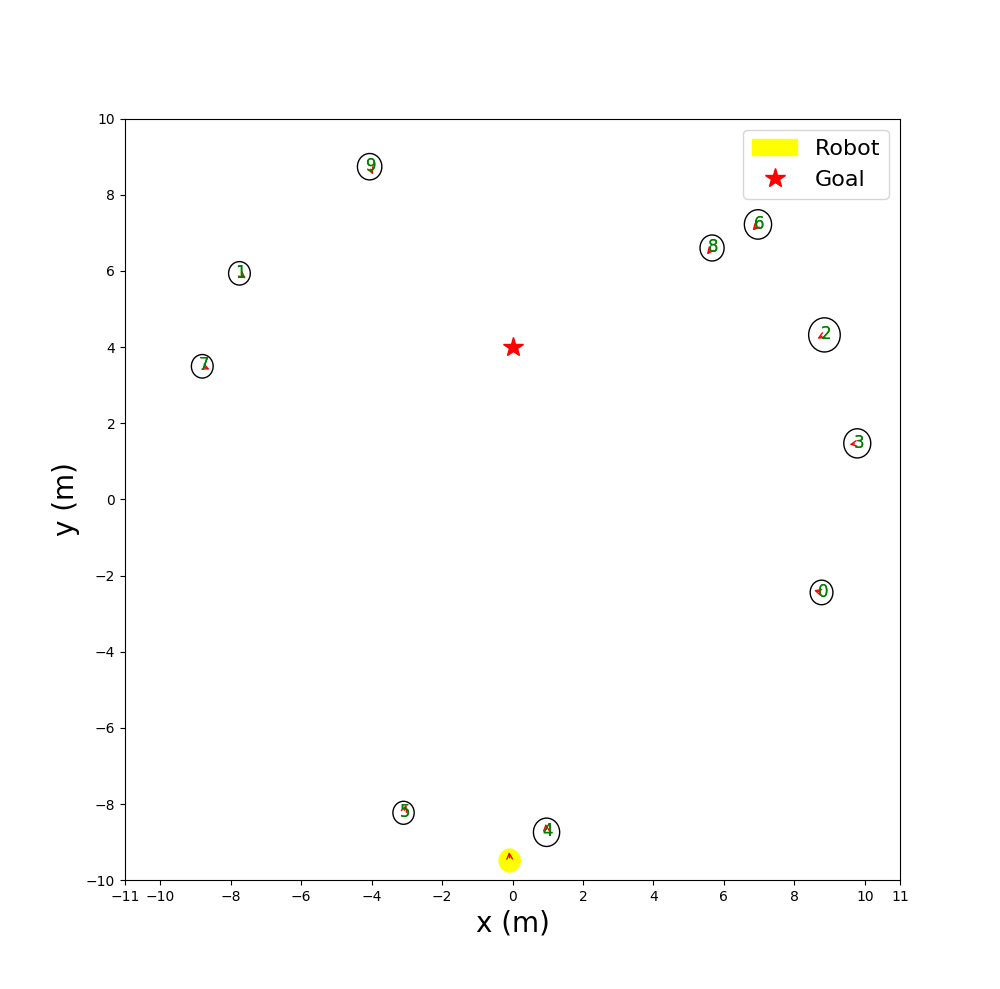}}
\subfloat[FOV simulator]{\includegraphics[width=3.5CM]{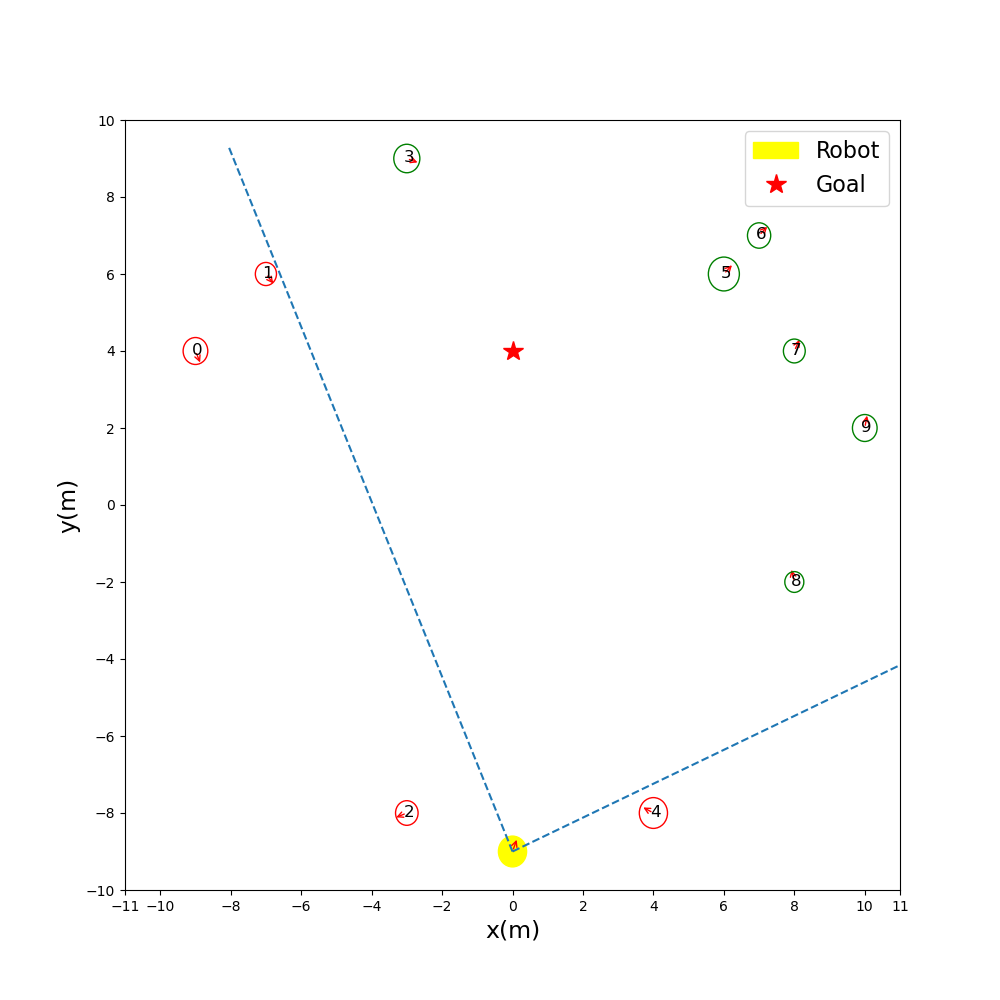}}
\caption{An illustration of the simulator: (a) We designed a gym-based simulator in the experiment section, which is a $22~m\times20~m$ open space. In the simulator, the robot is represented by a yellow circle, and humans are abstracted by white circles with different social radii. The goal position of the robot is denoted by a red star; (b) The robot's FOV is shown in a similar environment, which displays observable agents as green circles and unobserved agents as red circles.}
\vspace{-10pt}
\label{fig33}
\end{figure}


\subsubsection{Baseline and Ablation Models}
To ensure fair comparison, we selected several baseline methods: ORCA as the traditional baseline, CADRL  \cite{chen2017decentralized} and SARL  \cite{chen2019crowd} as the deep V-learning baseline, RGL \cite{chen2020relational} as the model-based RL baseline, and SRNN \cite{liu2021decentralized} as the ST-graph algorithm baseline. To eliminate the effect of RL parameters, we implemented two ablation models. The first ablation model, called NaviSRNN, uses the preference learning and SAC training procedure based on the SRNN interaction network to evaluate the differences in the interaction network under the same conditions. The second ablation model, called STAR+PPO, employs a handcrafted reward function as described in \cite{chen2017decentralized} and the training procedure from \cite{liu2021decentralized}. We did not use FAPL \cite{wang2022feedback} as a baseline because its interaction network is similar to our ablation model.

\subsubsection{Training Details}
All the baseline algorithms were trained following their original papers and initial configurations. Subsequently, we tested all the policies in the same environment. Additinoally, we applied the same configuration to each ablation model, using a learning rate of $4\times {10}^{-5}$ and training for $1\times {10}^{4}$ episodes.

\subsubsection{Evaluation Approaches}
As shown in Fig. \ref{fig5}, we utilize two kinds of evaluation methods. The first one is a successful rate, which collects the number of successful cases from total of 500 test cases. And we designed an evaluation function (social score $\rm F_{\rm{SC}}$) to estimate the comprehensive performance of social navigation, which considers the navigation time, the dangerous segments percentage, and the uncomfortable level of the robotic path as follows:
\begin{equation}
\left.{\rm{F}}_{\rm{{SC}}}=100 \cdot\left[{\rm{v}} \cdot \rm{F}_{\text {time }}+(1-v\right) \cdot \rm{F}_{\text{scc}}+\rm{v}^\prime \cdot \rm{F F}\right]
\label{social score}
\end{equation}
\noindent where $\rm v\in[0,1]$ is a weighted parameter, which defines the relative rate of navigation time cost $(\rm{F}_{\text{time}})$ and social compliance cost $(\rm{F}_{\text{scc}})$, and $\rm v^\prime \in (\infty,0]$ is a penalty factor, which represents the importance of failure rate $\rm FF$ that is the collision and overtime rate for total cases. 

Final, the social score is normalized as $\rm F_{\rm{SC}} \in (-\infty,100]$. In our experiments, the $\rm v$ is 0.35 and $\rm v^{\prime}$ is 0.25. The navigation time cost function $\rm F_{\rm{time}}$ 
is defined as follows:
\vspace{-5pt}
\begin{equation}
    {\rm{F}}_{\rm{time}} = 1-\frac{1}{\rm{K_1}}\cdot\sum_{\rm k_1=1}^{\rm K_1}\frac{\rm t_{k_1}-\min}{\max-\min}   
\end{equation}
\noindent where ${\rm{K_1}}$ is the number of successful cases in the test, $\rm{t_{k_1}}$ is the navigation time of $\rm k_1$-th successful paths, $\min$ is the minimum cost of time in total cases, and $\max$ is the maximum time of successful paths. We normalize the $\rm F_{\rm{time}}\in [0,1]$ and define the social compliance cost $(\rm{F}_{\text{scc}})$ as follows:
\vspace{-5pt}
\begin{equation}
\rm{F}_{\text{scc}}=1-\frac{1}{2} \left[\frac{1}{K_2} \cdot \sum_{k_2=1}^{K_2} \operatorname{sigmoid} \left(\frac{d u \cdot \Delta t}{\int^{T} d i s \cdot d t} -1\right)+\frac{K_1}{K_2}\right]
\end{equation}
\begin{figure}[!t]
\centering
\includegraphics[width=1\columnwidth]{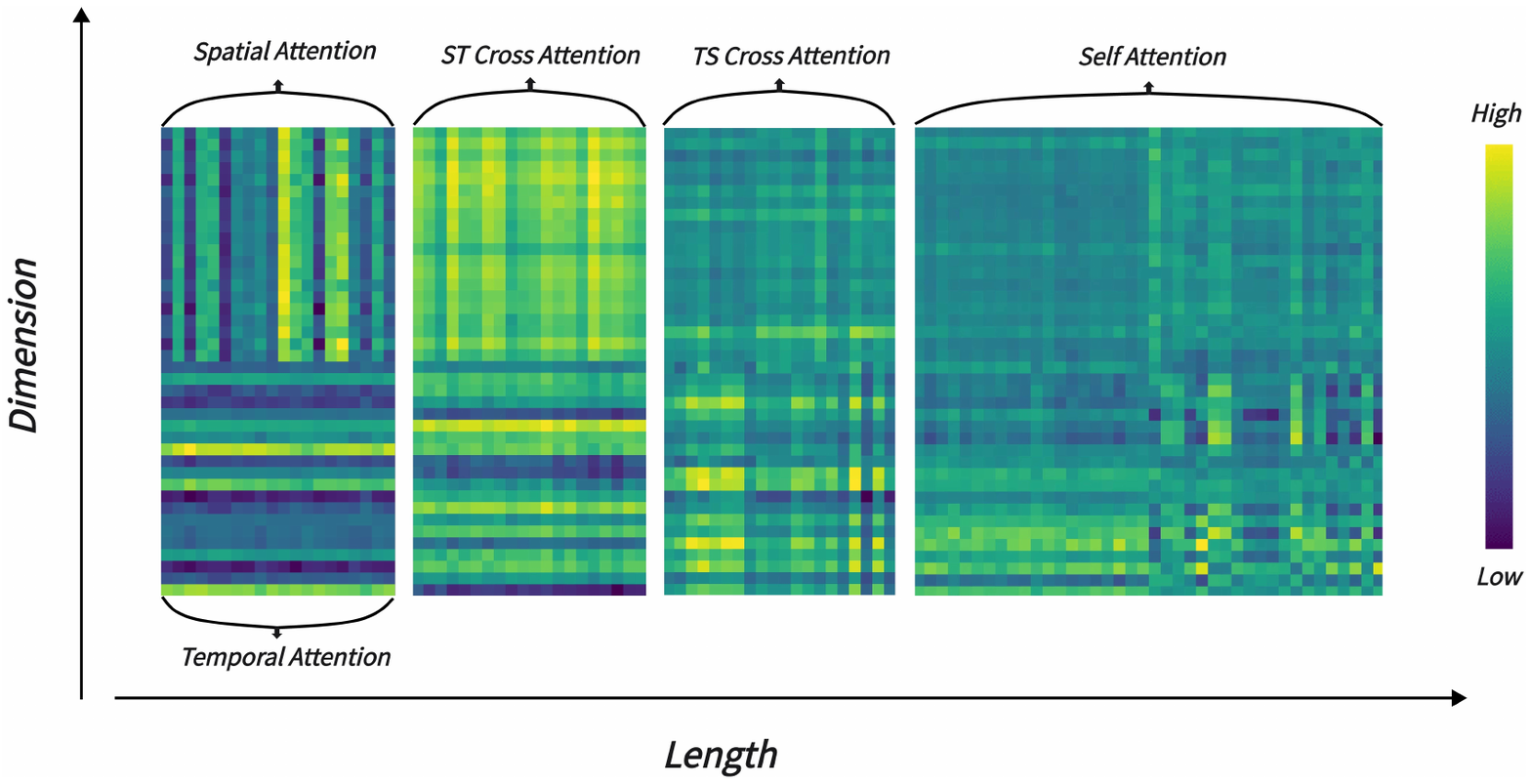}
\vspace{-10pt}
\caption{Attention visualization: Visualization of spatial-temporal attention matrices, cross-modal attention matrices and self-attention matrix from NaviSTAR.}
\vspace{-10pt}
\label{fig:attention martix}
\end{figure}
\noindent where ${\rm{K_2}}$ is the number of instances who are involved with uncomfortable segments, and ${\rm{du}}$ is the uncomfortable distance which is set as 0.45m in our environment from \cite{rios2015proxemics}, and ${\rm{dis}}$ is the minimization distance between robot and each pedestrian. The integral is a weighted average of robot ${\rm{dis}}$ from all cases in timestep $\rm t$. Finally, we normalize the $\rm F_{\rm{scc}}\in [0,1]$ by a standard nonlinear function $\operatorname{sigmoid}(\cdot)$ to evaluate the average impact caused by dangerous segments.
\subsection{Qualitative  Analysis}
\subsubsection{Overall}
NaviSTAR demonstrated a better performance due to its powerful interaction neural network, which can implicitly encode deeper potential interactions and cooperation among humans and robot. As shown in Fig.~\ref{fig:attention map}, NaviSTAR aggregates all the spatial and temporal attention maps from each individual agent. Fig.~\ref{fig:attention martix} shows the visualization of attention matrices where each cell represents an attention score between two feature blocks. The spatial attention matrix shows the relative importance of each agent to its surroundings in each timestep, while the temporal matrix presents the importance of all agents' self-trajectory states. The cross-attention matrix and self-attention matrix abstract the results of the fusion to capture environmental dynamics.

We compared the success rate and social score of all algorithms under the same configuration, which includes 10 humans and 18 $m$ of target distance, in an open space with FOV-360° and FOV-90° conditions separately. The experiments were tested by 500 random cases, as shown in Fig.~\ref{fig5} Table~\ref{table_1}.


\begin{table}[h]
\vspace{-5pt}
\caption{Social Score Table\label{tab:table1}}
\centering
\begin{scriptsize}
\begin{tabular}{cccccccc}
\hline & \multicolumn{3}{c}{ Social Score } & & \multicolumn{3}{c}{ Social Score } \\
\cline { 2 - 4 } \cline { 6 - 8 } Methods & \multicolumn{3}{c}{ FOV } & & \multicolumn{3}{c}{ Human Number } \\
& 90 & 180 & 360 & & 10 & 15 & 20 \\
\hline ORCA \cite{ORCA}& $21.1$ & $23.2$ & $27.4$ & & $29.2$ & $25.7$ & $22.6$ \\
CADRL \cite{chen2017decentralized}& $62.7$ & $65.4$ & $69.6$ & & $77.1$ & $73.2$ & $70.1$ \\
SARL \cite{chen2019crowd}& $31.3$ & $34.1$ & $39.7$ & & $42.2$ & $39.1$ & $36.7$ \\
RGL \cite{chen2020relational}& $76.5$ & $81.3$ & $84.9$ & & $90.3$ & $88.3$ & $86.6$ \\
SRNN \cite{liu2021decentralized}& $80.4$ & $84.7$ & $88.6$ & & $87.1$ & $84.8$ & $81.2$ \\
STAR+PPO & $85.1$ & $88.9$ & $91.3$ & & $\mathbf{99.2}$ & $95.4$ & $94.3$\\
NaviSRNN & $89.3$ & $91.6$ & $\mathbf{94.5}$ & & $95.7$ & $93.4$ & $91.2$\\
NaviSTAR   & $\mathbf{91.7}$ & $\mathbf{93.1}$ & $94.2$ & & $98.9$ & $\mathbf{98.5}$ & $\mathbf{95.8}$\\
\hline
\end{tabular}
\end{scriptsize}
\vspace{-5pt}
\label{table_1}
\end{table}

\subsubsection{Traditional Methods}
As shown in Fig.~ \ref{fig5} and Table~ \ref{tab:table1}, the traditional methods cannot be well deployed in a crowded environment with the assumption of the invisible robot. In our tests, ORCA showed the lowest success rate by a one-step avoidance policy in a dynamic environment due to strategic short-sight. Except for the failure cases, ORCA also represented a bad social performance, since the robot could not understand complex interactions. 

\subsubsection{Learning-based Methods}
In Fig.~\ref{fig5} and Table~\ref{table_1}, CADRL and SARL, as previous learning-based socially aware navigation approaches, exhibited limited performance and social compliance because they cannot capture deeper and full HRI. Secondly, RGL did not demonstrate a good performance as well, possibly due to the lack of temporal feature representation of RGL and model-based RL method's complex parameter engineering requirement. Thirdly, compared to other baseline algorithms, SRNN demonstrated a better HRI reasoning ability using an ST-graph and recurrence framework that can describe the spatial and temporal relationship of the crowd. However, due to the limitations of the handcrafted reward function, SRNN did not present a satisfactory social norms compliance.
\begin{figure}[!t]
\centering
\vspace{-5pt}
    \begin{subfigure}[b]{0.22\textwidth}
    \centering
       \includegraphics[width=\textwidth]{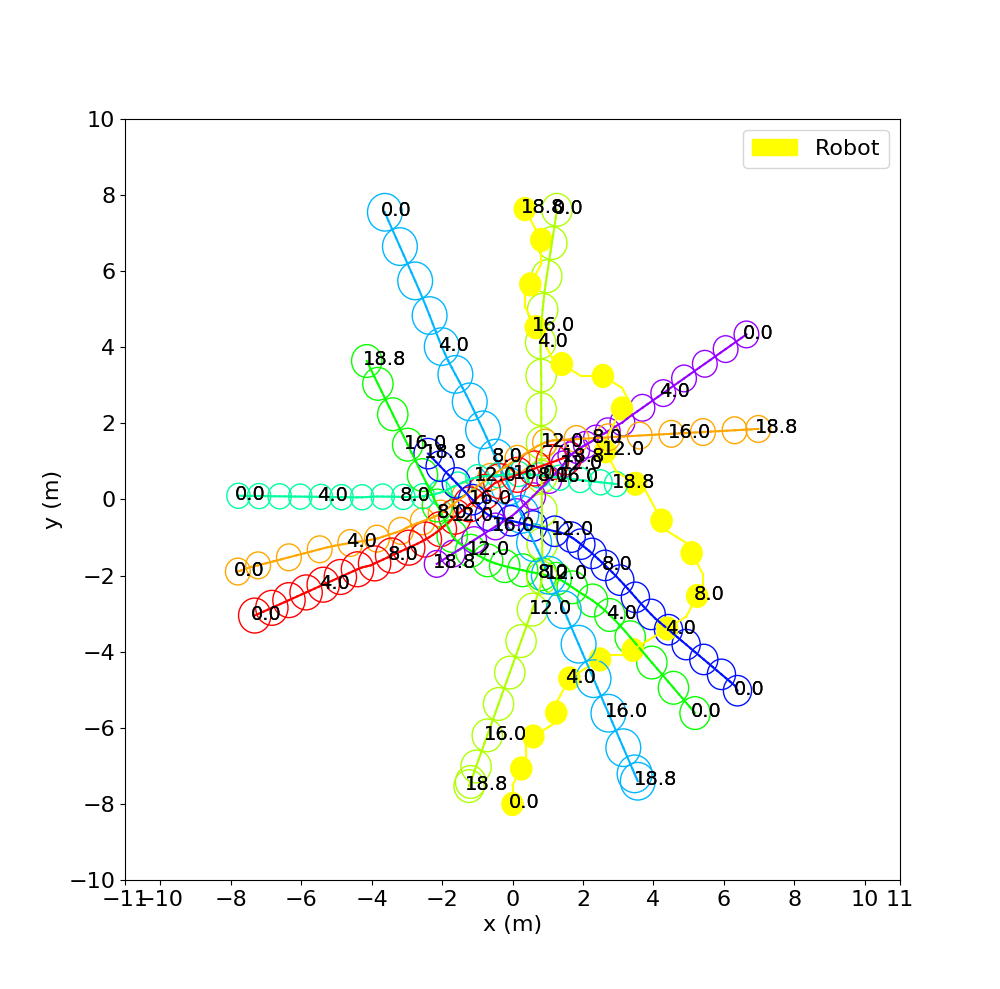}
       \vspace{-20pt}
        \caption{Policy: STAR+PPO}
        \label{fig:c}
    \end{subfigure}
    \begin{subfigure}[b]{0.22\textwidth}
    \centering
        \includegraphics[width=\textwidth]{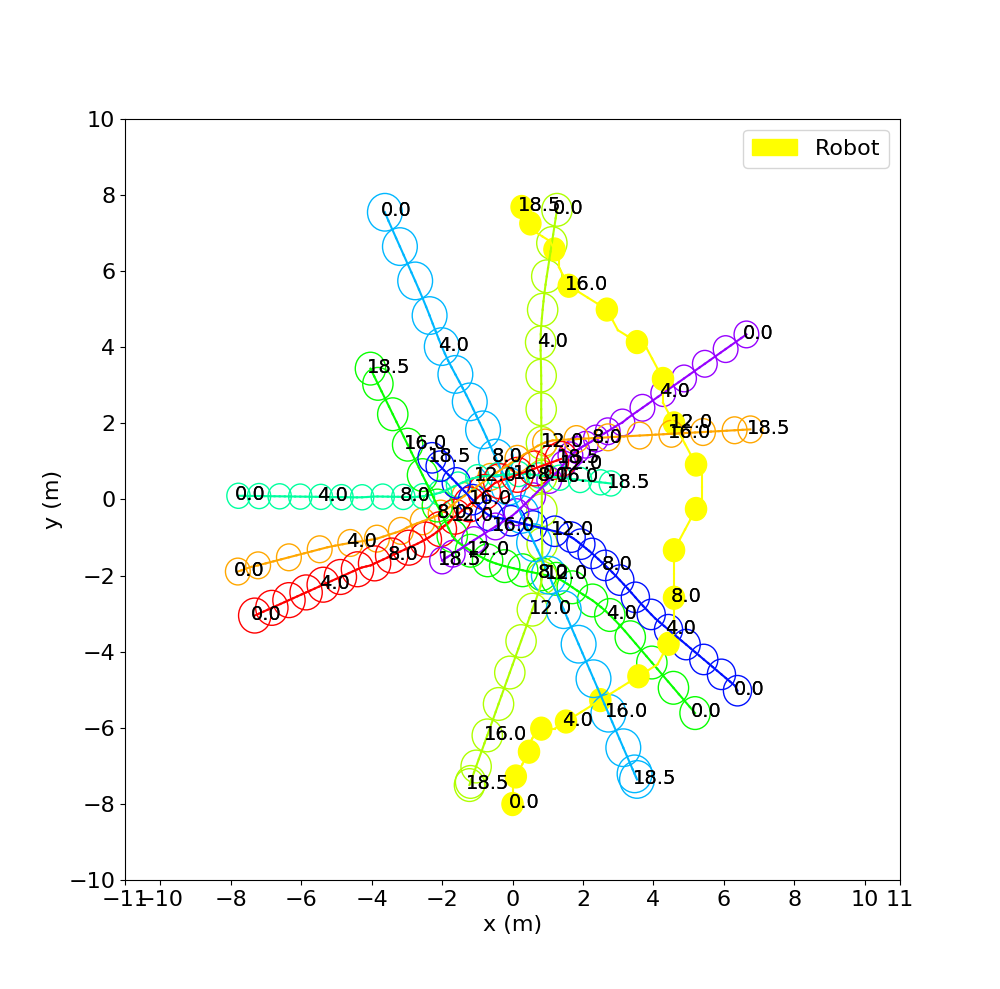}
        \vspace{-20pt}
        \caption{Policy: NaviSTAR}
        \label{fig:d}
    \end{subfigure}\\
    
        \begin{subfigure}[b]{0.22\textwidth}
    \centering
       \includegraphics[width=\textwidth]{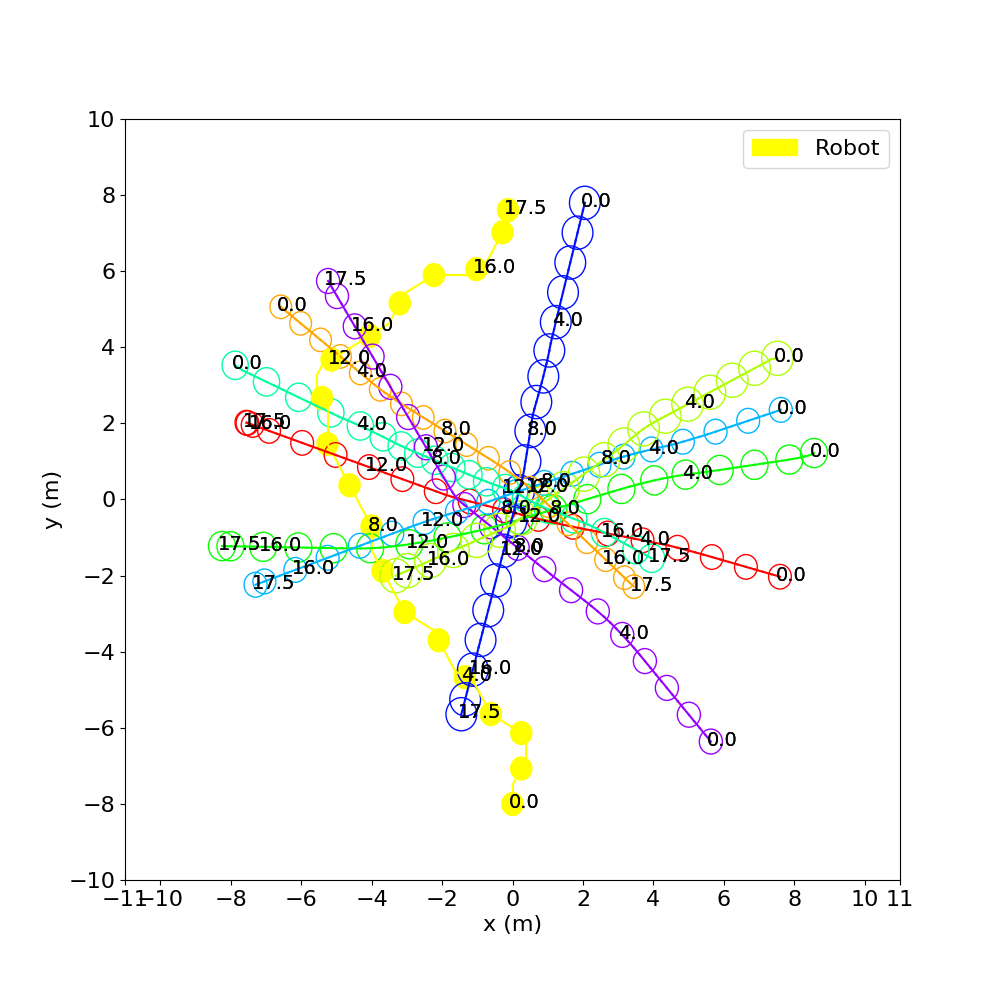}
       \vspace{-20pt}
        \caption{Policy: NaviSRNN}
        \label{fig:g}
    \end{subfigure}
    \begin{subfigure}[b]{0.22\textwidth}
    \centering
        \includegraphics[width=\textwidth]{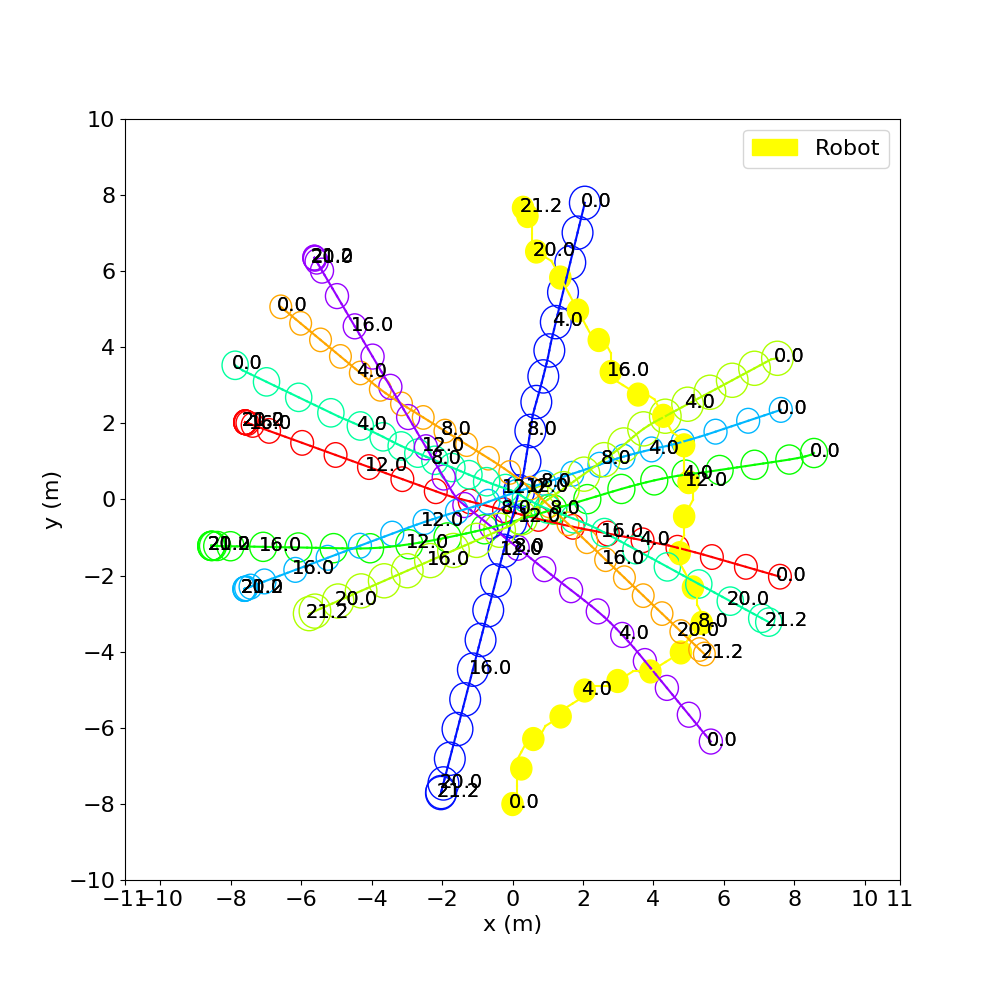}
        \vspace{-20pt}
        \caption{Policy: NaviSTAR}
        \label{fig:h}
    \end{subfigure}
    \caption{Simulation tests with ablation models and NaviSTAR: The trajectories of the first row are tested by a same case in open space, and the second row is tested by another case in FOV-90°. The yellow circle is robot trajectory, and other hollow circles are human trajectories.}
    \vspace{-10pt}
    \label{fig3}
\end{figure}

\subsubsection{Ablation Models}
Based on the trajectories shown in Fig. \ref{fig3}, we can observe that NaviSTAR tended to navigate through the side with fewer potential risks from crowd intents understanding, while NaviSRNN followed SRNN's previous path orientation with higher collision risks. This suggests that the spatio-temporal graph transformer used in NaviSTAR has better abilities to leverage long-term dependencies and more fully correlations compared to SRNN's network. Furthermore, NaviSTAR and STAR+PPO are good at interaction representation to navigate along a safer way, and NaviSTAR and NaviSRNN demonstrate a better social norm compliance than STAR+PPO. These findings suggest that incorporating a spatio-temporal graph transformer network and preference learning framework can significantly improve the performance and social compliance of socially aware navigation algorithms in human-filled environments.

\begin{figure}[!t]
\centering
\includegraphics[width=\columnwidth]{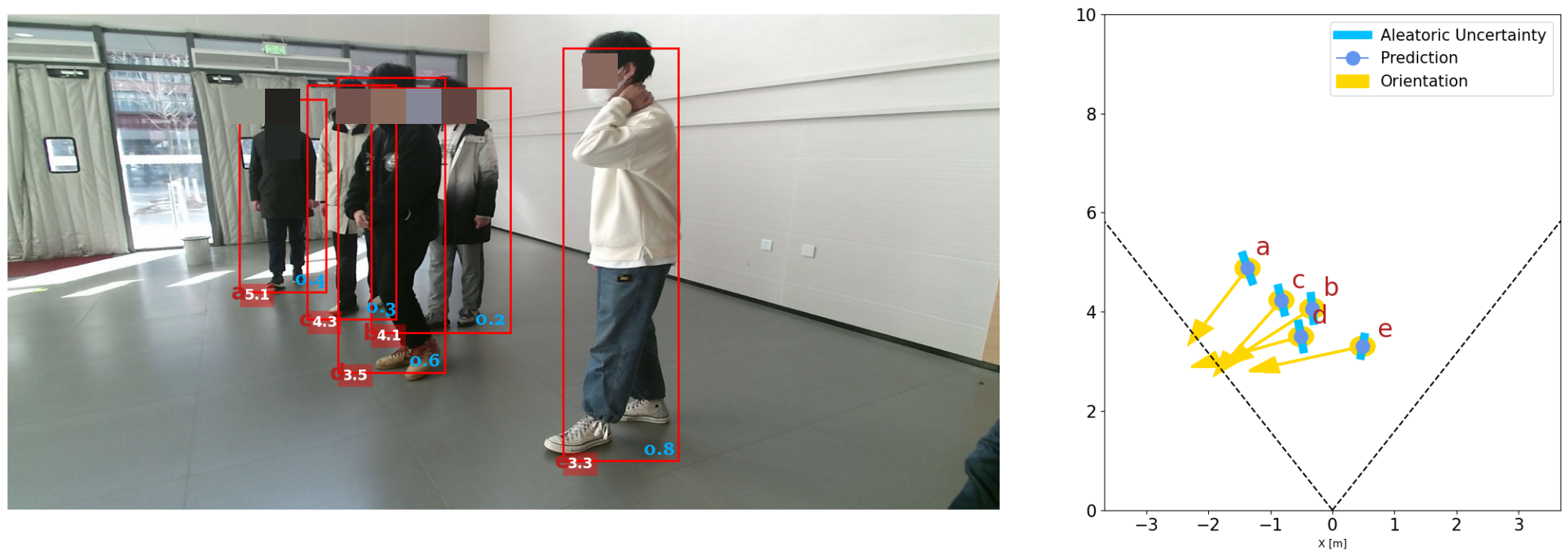}
\vspace{-15pt}
\caption{A robotic perception illustration: The velocity and distance predictors (left) present the distance between the robot and pedestrians in a red block and estimate the velocity of pedestrians with a blue number. The robotic 2D local map (right) shows the relative localization of pedestrians with velocity direction. An objective detector is deployed to capture the labels of users.}
\vspace{-10pt}
\label{fig7}
\end{figure}

\subsection{Real-world Experiment}
We also conducted real-world tests with 8 participants (1 female and 7 males, all aged over 18)\footnote{This user study was reviewed by the BUCT Institutional Review Board (IRB)}. During the tests, both the robot and participants adhered to the same behavior policy, starting point, and end goal for each case. The planner of the robot was unknown and randomized for each participant. Pre-experiment and post-experiment usability questionnaires were distributed to collect background data and feedback on the robot. To enable robotic perception, we developed a system that utilized a velocity predictor based on YOLO with Deepsort \cite{pramanik2021granulated} and a distance estimator from \cite{bertoni2021monstereo}, as shown in Fig. \ref{fig7}. The baselines selected for the user study were NaviSTAR, NaviSRNN, STAR+PPO, and SRNN. In the training procedures of above algorithms, over 10 human supervisors selected 5,000 preferences of segments. During the testing phase, we deployed a robot with Kinect V2, providing an 86° FOV. All algorithms were tested with the same configuration, and each algorithm was tested five times. 

The post-questionnaire evaluated real-world tests using the comfort index and naturalness index from users' feedback, based on \cite{wang2022feedback}. The responses ranged from strongly disagree (1) to strongly agree (5). We collected feedback scores based on participants' experiences interacting with the robot, where the two index factors were summed together and then multiplied by $10$ to map the results to the range $[20,100]$. According to the feedback received, NaviSTAR achieved an average score of 93, which was higher than NaviSRNN's score of 89, STAR+PPO's score of 86, and SRNN's score of 80. More detailed information about this user study setting and results can be found on our $\text{website}^{\ref{2}}$.
\vspace{-5pt}
\section{Conclusion}
In this paper, we proposed NaviSTAR, a novel benchmark for socially aware navigation that leverages a cross-hybrid spatial-temporal transformer network to understand crowd interactions with preference learning to learn social norms and human expectations. We supplemented equations-(17,18,19) to develop FAPL \cite{wang2022feedback} as NaviSTAR's training procedure. Additionally, we have also designed a new evaluation function ${\rm{F}}_{\rm{{SC}}}$ for socially aware navigation tasks. Through extensive experiments in both simulator and real-world scenarios, we demonstrated that NaviSTAR outperforms previous methods. We believe that our algorithm has the potentiality to make people feel more comfortable with robots navigating in shared environments. 

\vspace{-5pt}
\bibliography{main}
\bibliographystyle{IEEEtran}
\end{document}